\documentclass[12pt, final]{arxiv}

\title[Zero-Shot Robot Manipulation from Passive Human Videos]{Zero-Shot Robot Manipulation from Passive Human Videos}

\usepackage{times}
 \usepackage{booktabs}
 \usepackage{wrapfig}
 \usepackage{amsfonts}       
\usepackage{nicefrac}       
\usepackage{microtype}     
\usepackage{amsmath}
\usepackage{graphicx}

\author{%
 \Name{Homanga Bharadhwaj$^{1,2}$, Abhinav Gupta$^{1}$, Shubham Tulsiani$^{1,*}$,Vikash Kumar$^{2,*}$ } \\
 \addr{$^{1}$ The Robotics Institute, Carnegie Mellon University}\\
 \addr{$^{2}$Meta AI}\\
 \addr{$^*$Equal Contribution}
}

\newcommand{\VK}[1]{} 
\newcommand{\HB}[1]{} 
\newcommand{\shubham}[1]{} 

\begin{document}

\maketitle

\begin{abstract}%
Can we learn robot manipulation for everyday tasks, only by watching videos of humans doing arbitrary tasks in different unstructured settings? Unlike widely adopted strategies of learning task-specific behaviors or direct imitation of a human video, we develop a a framework for extracting agent-agnostic action representations from human videos, and then map it to the agent's embodiment during deployment. Our framework is based on predicting \textit{plausible} human hand trajectories given an initial image of a scene. After training this prediction model on a diverse set of human videos from the internet, we deploy the trained model \textit{zero-shot} for physical robot manipulation tasks, after appropriate transformations to the robot's embodiment. This simple strategy lets us solve coarse manipulation tasks like opening and closing drawers, pushing, and tool use, without access to \textit{any} in-domain robot manipulation trajectories. Our real-world deployment results establish a strong baseline for action prediction information that can be acquired from diverse arbitrary videos of human activities, and be useful for zero-shot robotic manipulation in unseen scenes. \footnote{\texttt{hbharadh@cs.cmu.edu} Videos  \url{https://sites.google.com/view/human-0shot-robot}}
\end{abstract}

\begin{keywords}%
  Learning from human videos, robot manipulation
\end{keywords}

\section{Introduction}

We humans effortlessly perform a plethora of manipulation tasks in our everyday lives, for example opening cabinets, cutting vegetables, pouring coffee, turning knobs, etc.  A common goal in the rapidly growing area of (data-driven) robot learning is to develop agents that can similarly perform diverse tasks in unstructured settings. Deep reinforcement learning based methods~\cite{mtopt} provide a framework that allows robots to continually improve at performing generic tasks by optimizing a corresponding reward. However, the sample (in)efficiency, the need for online interactions, and the difficulty in designing rewards and environment resets typically narrows their application to specific tasks in structured environments. An alternate approach is to directly learn action policies from robot demonstrations with experts~\cite{Pomerleau1989,visual_imitation1}. While demonstrations across varied settings and tasks can in principle allow learning the desired diverse behaviors, these are typically collected either through tele-operation or kinesthetic teaching and are thus difficult to scale. Moreover, both the online interactions and the robot demonstrations used in these approaches apriori require task definitions, and are restricted to lab settings with little variation, thus being a far cry from the diverse data required to train generalist robots effective in unknown unstructured environments.

\begin{figure}
    \centering
    \includegraphics[width=\textwidth]{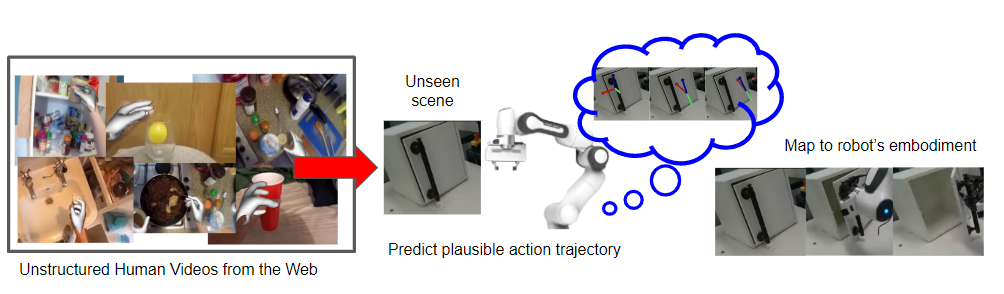}
\caption{\footnotesize Overview of the proposed framework for zero-shot robot manipulation from passive human videos. Our approach is based on hand trajectory prediction given an image of a scene, from human videos on the web, and transforming the predictions for zero-shot robot manipulation given an unseen scene in the lab.}
  \label{fig:teaser}
\end{figure}

Is there an alternate source of data that can enable learning for robot manipulation?
In this work, we show that  videos of humans interacting with objects as they accomplish everyday tasks serve as a readily available source of such large-scale data. Specifically, we show that  modeling the observed motion of the human hands across  human-object interaction videos allows \emph{zero-shot} prediction of the actions a robot should take to achieve goals in its environment. 
We develop a framework where given an initial image, the model learns to predict the sequence of motions of a human hand acting in the scene in subsequent frames. In particular, the predictions involve predicting a 6D pose of a point on the center of the palm of the human hand, over a 2 second future horizon from the initial image. Since there are several different plausible actions possible in scene, we make the model stochastic, and train it with diverse egocentric videos from existing datasets like Epic-Kitchens~\cite{epic}. Due to the scale and diversity of the human videos, the prediction model allows us to perform zero-shot action prediction in different physical environments with novel objects. We call such human videos \textit{passive} as they haven't been actively collected for the purpose of robot learning with any task-level segregation, and our framework \textit{zero-shot} because there are no in-lab demonstrations or fine-tuning. Fig.~\ref{fig:teaser} shows an overview of our framework for zero-Shot robot manipulation from passive Human videos 

We develop two versions of the framework: un-conditioned, and goal-conditioned. The un-conditioned model predicts plausible trajectories given a scene, and corresponds to realizing the different affordances of objects in a scene that can be helpful for downstream exploration. The goal-conditioned model is further prompted with a goal image to predict trajectories that reach the specified goal, and is helpful in achieving targeted behavior. To manipulate a robot with the predictions, we transform them to the end-effector space of the robot, and use a simple IK controller to execute the motion of the trajectory. We show that this simple approach works reliably well with around 40-60\% success rate for unconditionally manipulating objects like toasters, drawers, bowls etc. (see Fig.~\ref{fig:scene_objects}) and with around 30-40\% success rate for goal-conditioned manipulation. Finally, we note that incorporating some kind of fine-tuning with environment-specific data would definitely lead to better performance, for fine-grained manipulation tasks, but our current framework is a surprisingly strong baseline of what can be done zero-shot from robot-agnostic human data.

\section{Related Works}
We discuss prior works that attempt understanding of human activities in videos through detection of hand-poses, semantic tasks, visual feature extraction, and activity forecasting. We follow this with a discussion of papers that learn robot manipulation skills from videos of humans and robots performing different activities. \\

\noindent\textbf{Scaling Human-Object Interaction Understanding.} Understanding human activities has recently received a lot of interest with development of large-scale datasets~\cite{smthsmth,youcook,cmudata,ego4d} that involve recording videos of humans doing cooking related activities in their kitchen~\cite{epic,egtea}, short clips involving manipulating objects~\cite{smthsmth}, and more diverse and long clips of activities both inside and outside homes~\cite{ego4d,Shan20}. 

Based on these large datasets of human videos, several prior works have focused on understanding human-object interactions~\cite{Shan20,ye2022s}. Specifically, prior work has investigated object pose estimation~\cite{Kehl2017SSD6DMR,Rad2017BB8AS,Xiang2018PoseCNNAC,Hu2019SegmentationDriven6O,He2020PVN3DAD}, hand pose estimation~\cite{zimmermann2017learning,iqbal2018hand,spurr2018cvpr,ge20193d,baek2019pushing,boukhayma20193d,hasson2019learning,dkulon2020cvpr,liu2021semi}, full body pose estimation~\cite{rong2020frankmocap}, and hand-object joint pose estimation~\cite{ye2022s,GarciaHernando2018FirstPersonHA,hampali2020honnotate,liu2021semi,chao2021dexycb}. These efforts have been aptly complemented by datasets of 3D scans of real objects -- like YCB~\cite{calli2015benchmarking}, and Google Obbject Scans~\cite{downs2022google} that humans typically interact with. Another line of work has investigated learning interaction hotspots~\cite{nagarajan2019grounded,liu2022joint,goyal2022human} from videos, and predicting plausible grasps~\cite{mo2021where2act,brahmbhatt2019contactgrasp}. 

Building upon these developments, which focused primarily on visual understanding, our work focuses on \textit{closing the vision-robotics loop} by using large passive datasets of human videos. We learn plausible hand trajectories for interaction with objects, and deploy them for real robot manipulation. In the next sub-section we outline how our framework differs from robot learning approaches that learn manipulation skills from videos.\\

\noindent\textbf{Robot Manipulation from Videos}  
 Recent developments in robotics has extensively focused on using increasingly unstructured data. Visual imitation learning approaches aim to learn control policies from datasets of visual observations and robot actions~\cite{visual_imitation1,visual_imitation2,visual_imitation3}. Behavior cloning~\cite{Pomerleau1989,Bain1995,ross2010efficient,Bojarski2016,Torabi2018b} and inverse reinforcement learning ~\cite{Russell1998,Ng2000,Abbeel2004,Ho2016,Fu2017,Aytar2018,Torabi2018g,Liu2020} are two popular approaches in this regime, but are difficult to scale to unseen scenes as they require high quality in-domain expert robot trajectories  (typically) with a human controlling the robot. 
 To alleviate the need for collecting high quality robot data, some approaches have used videos of humans doing things~\cite{schmeckpeper2019learning,chang2020semantic,schmeckpeper2020reinforcement,shawvideodex,shao2020concept,song2020grasping,young2020visual} for learning control policies either through imitation of reinforcement learning. However, for imitation, data needs to be collected through special hardware interfaces by humans in different scenes for visual imitation~\cite{young2020visual,song2020grasping} which is hard to scale.  For the RL based approaches, the frameworks require simulation environments for online interactions with resets and rewards to provide feedback during learning~\cite{shao2020concept,schmeckpeper2020reinforcement}. Compared to these approaches, we do not require any specially collected data, or simulators for learning, thereby generalizing zero-shot to unknown tasks. 
 
 Other works have investigated learning robot motions through a direct imitation of a corresponding human video in the scene~\cite{Peng2018s,Pathak2018,Sharma2018,sieb2020graph,sivakumar2022robotic,garcia2020physics,xiong2021learning,Sermanet2018,sharma19thirdperson,smith2019avid,peng2020learning,bahl2022human,qin2021dexmv}. Some of these approaches~\cite{smith2019avid,xiong2021learning} although do not require large-scale datasets, requre near-perfect alignment in the poses of the robot and human arms. Recent works~\cite{bahl2022human} have alleviated the need for this perfect alignment, through in-painting of humans from the scene to construct a reward function, but they still require per-task online fine-tuning through RL, which is expensive in the real-world. DexMV~\cite{qin2021dexmv} combines a vision and simulation pipeline to effectively translate human videos to dextrous hand motions, but requires several in-lab human videos for training and cannot utilize existing human videos on the web.

Compared to these prior works, we extract agent-agnostic representations in the form of hand-trajectories from human videos on the internet. Instead of having a video stream to mimic directly, we learn a scene-conditioned trajectory prediction model that relies on a single image observation as input and can be directly used for robot manipulation zero-shot. This enables generalization of manipulation capabilities to unseen scenes, without apriori notions of task specification.

\section{Approach}
Motivated by the intuitive understanding humans have about how a scene can be manipulated in interesting ways, we develop a computational framework that can endow robots with similar manipulation skills, by only observing videos of humans interacting with objects in diverse unstructured settings. Specifically, given an initial image of a scene $o_1$ from a single viewpoint, we learn a model $p(a_{1:T}|o_1)$ that predicts future actions $ a_{1:T}$ taken by humans while interacting in the scene. We focus on table-top manipulation setting where $a_{1:T}$ corresponds to plausible 6D poses of a right human hand. In order to make the framework generally applicable for a robot arm with a simple end-effector like a  two-finger gripper, we do not model the hand articulation (i.e. how each joint is oriented with respect to the wrist), and focus on predicting only the position and orientation of the palm of the hand. After training this prediction model across diverse internet videos, we apply it to an in-lab setting with no additional fine-tuning on lab data, for manipulating objects like drawers, cabinets, and doors on a table top setting with a fixed camera.

\subsection{Learning to Predict Future Hand Trajectories}
Given an image of a scene, there could be several plausible trajectories that modify objects in the scene. To capture this multi-modality, we develop a stochastic model for $p(a_{1:T}|o_1)$ such that different samples from the model yield different trajectories. Instead of predicting absolute hand poses at each time-step $h_t$, we predict delta actions $a_t = h_{t} \ominus h_{t-1}$, where $\ominus$ is an appropriate ``difference operator." This minimizes error in predictions by reducing the absolute magnitude of model predictions. In addition, to enforce temporal correlation, the model is auto-regressive such that it takes in previous actions within a trajectory as input while predicting the next action. In the next sub-sections, we describe the prediction model architecture, and the pre-processing of data for training the model.

\subsubsection{Model Architecture}
\label{sec:model}

\begin{wrapfigure}[11]{R}{0.7\textwidth}
    \centering
    \vspace{-2em}
 \includegraphics[width=0.7\textwidth]{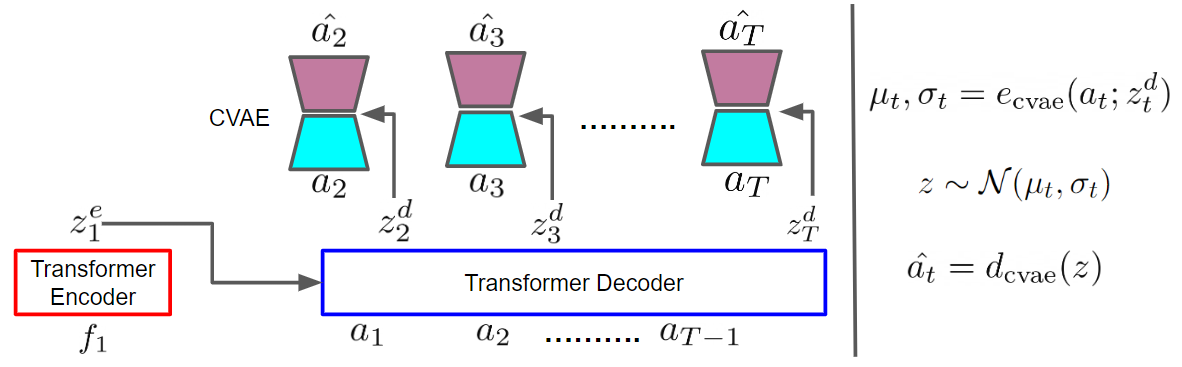}
    \vspace{-1.5em}
    \caption{\footnotesize Architecture of the hand trajectory prediction model showing the unconditioned trajectory prediction model given features of the initial scene. On the right we show the process of inference from the model. Details are mentioned in section~\ref{sec:model}}
    \label{fig:architecture}
\end{wrapfigure} 

The hand prediction model is an Image Transformer~\cite{bertasius2021space,chen2020generative}, with an Encoder-Decoder architecture, and a final C-VAE hand trajectory prediction head conditioned on the decoder feature outputs. We show an overview of the architecture in Fig.~\ref{fig:architecture}. Similar to prior works~\cite{rulstm,wang2016temporal}, we first extract features offline from the initial image $o_1$, and denote it as $f_1$. The feature $f_1$ is encoded by the transformer encoder $\mathcal{E}$ into a latent code $z^e_1$. The transformer decoder $\mathcal{D}$ conditions on the latent code $z^e_1$ and the previous actions $a_{1:t-1}$ to output the feature for action prediction at time-step $t$, $z^d_t$ . 

During training, a CVAE takes as input the action $a_t$ and conditional context $z^d_t$, and outputs reconstruction $\hat{a_t}$. During inference, we simply sample from the prior of this CVAE, concatenated with the predicted context $z^d_t$, and obtain the sampled delta action $\hat{a_t}$. After obtaining the delta actions we can recover the predicted actions as $\hat{a_t} = \hat{a_{t-1}} \oplus \hat{a_t}$ $\forall t>1$, \VK{11/26:Check this. Should it be $\hat{h_{t-1}} \oplus \hat{a_t}$?} where $\oplus$ is an appropriate ``addition operator." For the goal-conditioned model, the Transformer encoder also takes in a goal image embedding $f_g$ with appropriate positional embedding, and outputs $z_g^e$. We mention specific details of the models in the Appendix.
\shubham{Notation suggestion: I don't like $a_t$ (which should be an 'action' as the variable for hand pose, and $\delta a_t$ then being called `delta action'. Can't we use $h_t$ and $a_t$ instead?)}\HB{done. I will double check for consistency throughout}

\noindent\textbf{Transformer.} The transformer encoder and decoder consist of several stacked blocks of operations that involve attention and MLP with LayerNorm. The decoder blocks have cross-attention with the query and value tokens being the encoded code $z^e_1$. Whereas, in the encoder the attention blocks are all self-attention. 

\noindent\textbf{CVAE.} The CVAE prediction head at each time-step consists of an encoder $e_{\text{cvae}}(\cdot)$ and a decoder $d_{\text{cvae}}(\cdot)$ neural networks. Conditioned on the predicted feature from the transformer decoder $z^d_t$, and the current delta action $a_t$ as input, the CVAE encoder outputs the mean $\mu_t$ and S.D. $\sigma_t$ of a Gaussian distribution. The CVAE decoder samples from this Gaussian and outputs a predicted delta action $\hat{ a_t}$. Formally,
$
    \mu_t,\sigma_t = e_{\text{cvae}}(a_t; z^d_t) \;\;\;\;\;\; z\sim\mathcal{N}(\mu_t,\sigma_t)\;\;\;\;\;\;  \hat{a_t} = d_{\text{cvae}}(z)
$

\noindent\textbf{Training loss.} The overall training loss is defined in terms of the output of the CVAE per-timestep aggregated over all time-steps $T$ in the prediction horizon. 
\begin{align*}
    \mathcal{L} = \sum_{t=2}^T\left[|| a_t - \hat{ a_t}||^2 - D_{\text{KL}}\left(\mathcal{N}(\mu_t,\sigma_t) ||\mathcal{N}(0,1)\right)\right]
\end{align*}
This overall loss is backpropagated through the entire prediction model that involves both the Transformer and the CVAE (through re-parameterization), and there is no intermediate supervision for any of the stages. This makes the approach generally applicable to in-the-wild human videos and reduces pre-processing overhead for training data, as described in the next section. In addition, the learned model is task-agnostic since there is no task-specific distinction in the human video clips, and given unseen scenes, it allows performing \textit{plausible} tasks zero-shot.

\subsubsection{Training Data Generation} 
We consider 2 second clips of egocentric videos from Epic-Kitchens that  involve people doing everyday household activities, especially in the kitchen, like cooking food, opening cabinets, moving objects from one location to another etc. For the goal-conditioned model, the last image of the clip corresponds to the goal that is input to the model. To obtain ground-truth poses of the hand in the frames within prediction window, we run hand-tracking with an off-the-shelf FrankMocap model~\cite{rong2020frankmocap}. FrankMocap outputs a weak-perspective camera $(t_x,t_y,s)$ and $(x,y,z)$ locations of all the hand joints, and orientation of the hand $(\alpha,\beta,\gamma)$ relative to a canonical hand. We only consider 
the center of the palm, and record its 6D pose relative to the predicted camera in the beginning of the 2 second window. In summary, the training data consists of pairs $\{(o_1,h_{1:T})\}$ where $o_1$ is the first image of a video clip and $h_t=(x_t,y_t,s_t,t_x,t_y,\alpha,\beta,\gamma)$ is the hand pose and camera parameters at future time $t$.  

\begin{figure}[t]
    \centering
    \includegraphics[width=\textwidth]{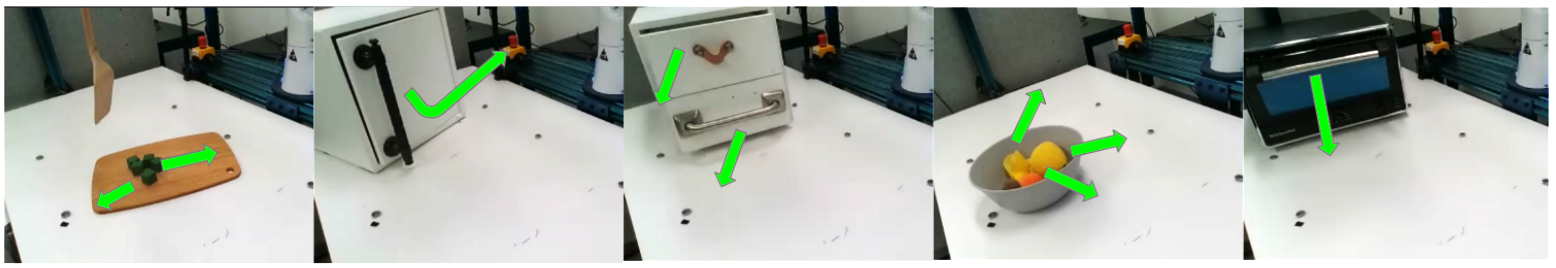}
    \caption{\footnotesize Figure showing a few configurations of the objects we place in the scene for our experiments, with green arrows denoting plausible motions; from left to right in the top row: a door with a vertical hinge, a bowl of fruits, a chopping board with veggies; in the bottom row: a toaster oven with horizontal hinge, a stack of two drawers, and a vegetable strainer. \VK{11/19: placement of object isn't something that will impress. What this figure needs to communicate is the diversity of tasks we can handle. Think how to communicate that instead. See task and dataset figure here -- https://sites.google.com/view/real-orl for motivation}\HB{Did you mean showing tasks through arrows?} \VK{11/26: Yes either through arrows or layered images with transparency that shows the tasks.}\HB{done}}
    \label{fig:scene_objects}
\end{figure}

\subsection{Mapping Trajectories to the Robot's Frame}
After training the overall hand pose prediction model, $p_\psi( a_{1:T}|o_1)$ with diverse internet videos, we deploy it directly for robot manipulation tasks in the lab. The robot sees an image of the scene through a fixed camera, and optionally receives a goal-image which is input to the prediction model. In order to use the actions predicted by the model $a_{1:T}$ for moving the robot, we need to transform them to the world coordinate frame of the robot, and considering each action $a_t$ as an end-effector target pose, use a low-level controller for executing the respective motions. 

 The camera in the scene is calibrated, so the intrinsic matrix $I$ and the extrinsic matrix $[R,T]$ are known. The world coordinates are located at the base of the robot (robot base is at same height as the table top) and the height of the table top from the camera is  known and approximately constant. Given scene from the camera $o_1$, the model predicts delta actions $a_{1:T}$ which we convert to absolute actions (described in section~\ref{sec:model}), and transform the actions from the camera frame to the world frame of the robot through inverse projection transformation. The prediction horizon is $T=7$
 for our experiments. After obtaining the world coordinates of the action sequence $\{(X_t,Y_t,Z_t,\alpha_t,\beta_t,\gamma_t)\}_{t=1}^T$, we use an IK controller to execute the corresponding motion for bringing the end-effector to the desired position and orientation and each time-step. We describe additional details in the Appendix.

\section{Experiment Design}
\begin{wrapfigure}[14]{R}{0.3\textwidth}
    \centering
    \vspace{-2em}
    \includegraphics[width=0.3\textwidth]{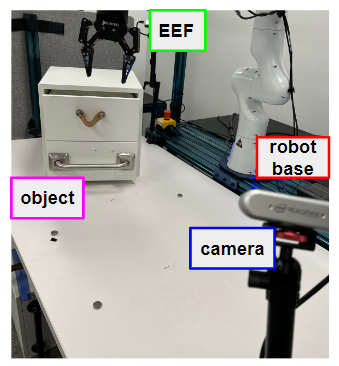}
    \vspace{-1em}
    \caption{\footnotesize Robot workspace showing the camera, an object in the scene, the Franka arm, and it's end-effector. \VK{11/26: The labels are blocking a lot of the image. Lets iterate on this figure. Also, improve caption}\HB{done}}
    \label{fig:workspace}
\end{wrapfigure} 
Through experiments, we aim to understand the following research questions:
\begin{itemize}
    \item In unknown tasks, how good is our unconditional model in generating diverse but plausible task outcomes?
    \item In known tasks, how good is our goal-conditional model and action mapping to solve tasks?
\end{itemize}

Qualitatively, we visualize the diversity and plausibility of trajectories executed for different scenes with the unconditioned model. For quantitative evaluations, we compare success rates of the both the unconditioned model predictions when the robot is placed in different scenes, and the goal-conditioned model when prompted in addition with a desired goal configuration of objects. We compare against a 3D scene flow~\cite{sceneflow,raft3d} baseline that uses RAFT3D~\cite{raft3d} for predicting scene flow field between the initial and goal images, and uses the dominant flow direction to guide the motion of the robot. To test the importance of training across diverse data, we compare against a version of our method that is trained on only 30\% of the total training data. We mention additional details of the setup, objects, tools, and comparisons for the experiments in the Appendix\footnote{\url{https://sites.google.com/view/human-0shot-robot}}.

\textbf{Experimental setup}: The workspace shown in Fig.~\ref{fig:workspace} is a white table with a camera in one of the corners, and the robot base on it's opposite edge. In Fig.~\ref{fig:scene_objects}, we show the different objects we place in the scene for our experiments. Note that all of the objects \VK{11/19: Lets talk in terms of tasks/activities, not objects}\HB{I'll take a stab at harmonizing, but am inclined to still talk largely in terms of objects placed in the scene, because we are pitching the framework as task-agnostic throughout} are unseen by definition because the egocentric videos used for training our models are from existing datasets on the web and do not involve any in-lab data.

\section{Results}
In this section, we discuss results for the un-conditional, and goal-conditional models followed by an analysis of failures. We show both qualitative results for visualization, and quantitative evaluations over several trials. 
\subsection{Un-conditional generation results}
\begin{wrapfigure}[13]{R}{0.5\textwidth}
    \centering
    \vspace{-1em}
    \includegraphics[width=0.5\textwidth]{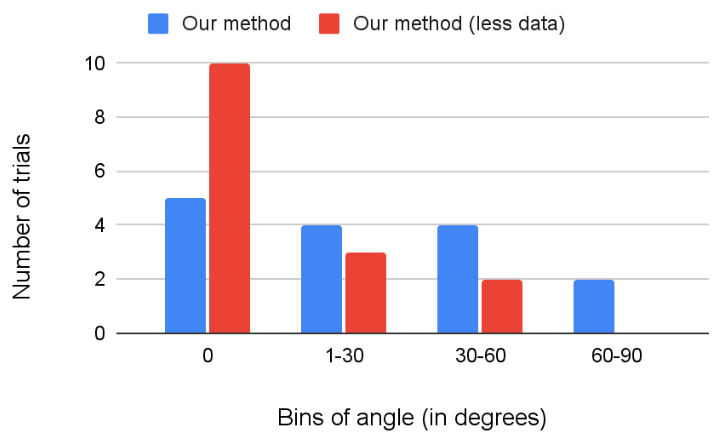}%
    \vspace{-0.8em}
    \caption{\footnotesize Fine-grained analysis of the door opening task. Histogram shows number of trials (out of 15) that open the door to a certain angle.}
    \label{fig:doorfine}
\end{wrapfigure}
\begin{figure}[h!]
    \centering
    \includegraphics[width=\textwidth]{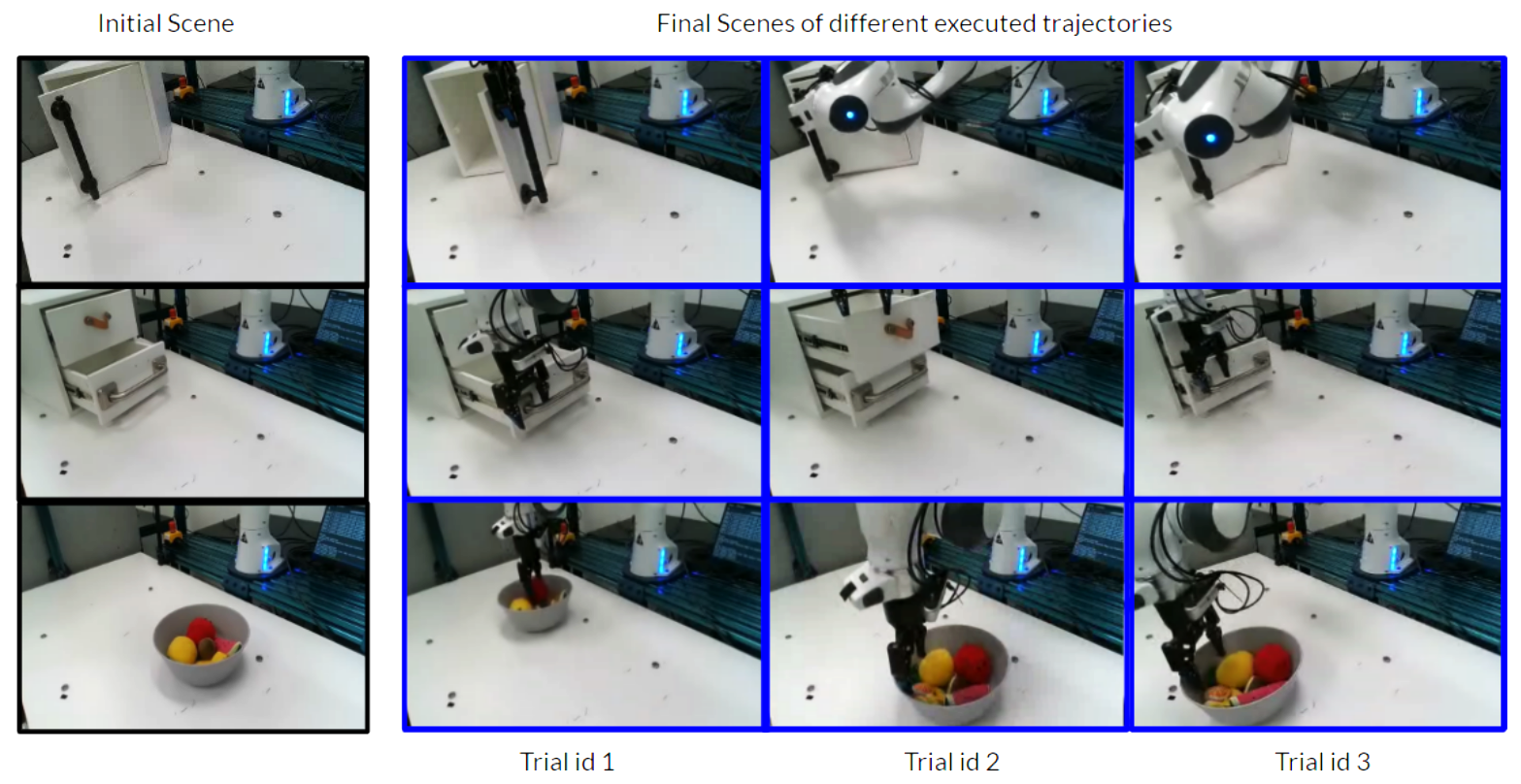}
    \caption{\footnotesize Figure showing final configurations corresponding to different executed trajectories (trial ids 1,2,3) by the unconditional model in the scenes, showing the diversity of plausible behaviors predicted by our model.}
    \label{fig:unconditional_qual}
    \vspace*{-0.5cm}
\end{figure}

 
For the un-conditional generations, given a scene, the model predicts a sequence of actions, which are executed by the robot. We evaluate the model in terms of whether the trajectories executed by the robot correspond to \textit{plausible} interactions in the scene that a human is likely to do. For example, a closed door can be opened, and a bowl of fruits can be moved around on the table i.e. given a scene, there is a distribution of plausible object state changes. Fig.~\ref{fig:unconditional_qual} shows visualizations of of predicted trajectories executed by the robot in different initial scenes. 

In Table~\ref{tb:unconditional} we analyze the unconditioned model generations quantitatively, for different objects in the scene with definitions of success criteria different for different objects. For a door, drawer, and toaster, success is when the robot changes objects' state - from open to close or vice versa. For the bowl, success is when the bowl is moved from an initial position to a different position without toppling. From the table we see that success rates vary between 45\% and 55\% indicating that the  model is able to predict trajectories that perform plausible interactions in a scene. Fig.~\ref{fig:doorfine}, we perform a finer analysis of the model predictions when a closed door is placed in the scene, and observe that over 70\% of the trials open the door to non-zero angle, indicating plausible state change of the object.

\begin{table}[t]
\centering
\caption{\footnotesize Quantitative evaluation results for the unconditioned model, with 20 trials for each object. For each object, we measure success differently. For a door, drawer, and toaster, success is when the robot opens the respective object from a closed initial position, or closes the respective object from an open initial position. For the bowl, success is when the bowl is moved from an initial position to a different position without toppling. For the moving of veggies task, success is when at least half the pieces on the table are moved.}
\begin{tabular}{@{}ccccccc@{}}
\toprule
                      & \textbf{Drawer} & \textbf{Door} & \textbf{Toaster} & \textbf{Bowl} & \textbf{Veggies} & \textbf{Average} \\ \midrule
\textbf{Our method} & 50\%            & 45\%          & 50\%             & 55\%       & 50\%     & 50\% \\
\textbf{Our method (less data)} & 10\%            & 10\%          & 15\%             & 20\%     & 15\%     & 14\%  \\
\bottomrule
\end{tabular}
\label{tb:unconditional}
\end{table}
\subsection{Goal-conditioned generation results}

\begin{figure}[t]
    \centering
    \includegraphics[width=\textwidth]{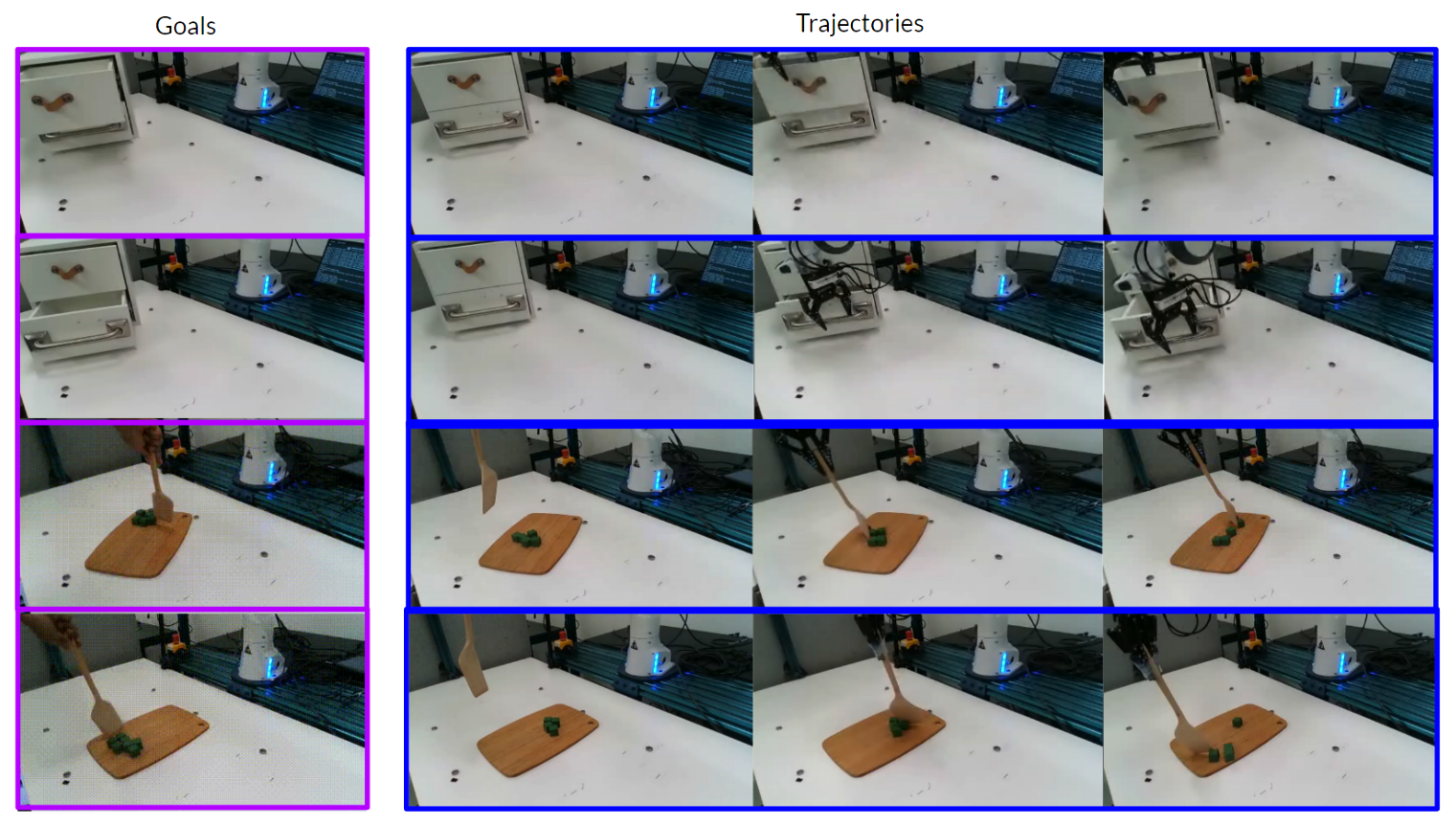}
    \caption{\footnotesize Figure showing different goal-conditioned robot evaluations for goals corresponding to the one shown on the left. Each row corresponds to a sub-sampled trajectory, respectively showing top and bottom drawers being opened, door being opened and closed, and veggies being moved on a table.}
    \label{fig:goalconditioned_qual}
    \vspace*{-0.7cm}
\end{figure}

In addition to executing plausible trajectories in a given scene, we want to understand how effective is the goal-conditioned model in generating trajectories that reach a specified goal from an initial scene. Fig.~\ref{fig:goalconditioned_qual} shows results for trajectories corresponding to different goal images. In Table~\ref{tb:goalconditional} we show quantitative results for different types of goal images. For each setting, we randomize the initial pose of the object, for example for a `open door' goal image, the initial position of the door is either closed, or is half-open. We see that the goal-conditioned model succeeds in predicting trajectories that succeed on an average around 37\% of the times, which is lower than that of the unconditioned model because the task of reaching a particular goal is more difficult than bringing unconditional changes to a scene. We observe better performance compared to the baselines trained on less data, and the scene flow baseline. In Table~\ref{tb:ungoal} we do a finer analysis of our model for a certain scene  with stacked drawers and observe that when conditioned on a goal, the goal-conditioned model reaches the final configuration specified by the goal more number of times than the unconditional model reaches the same. This demonstrates that goal-conditioning actually leads to predicted trajectories consistent with the specified goal.

\begin{table}[h!]
\vspace*{-0.5cm}
\centering
\caption{\footnotesize Quantitative evaluation results for the goal-conditioned model, with 20 trials for each goal category. Each column denotes the type of goal image specified. In the drawer object, we aggregate results for the top and bottom drawers in the opening and closing tasks, respectively.}
\begin{tabular}{@{}cccccccc@{}}
\toprule
                                                                & \textbf{\begin{tabular}[c]{@{}c@{}}Open \\ Drawer\end{tabular}} & \textbf{\begin{tabular}[c]{@{}c@{}}Open \\ Door\end{tabular}} & \textbf{\begin{tabular}[c]{@{}c@{}}Close \\ Drawer\end{tabular}} & \textbf{\begin{tabular}[c]{@{}c@{}}Close \\ Door\end{tabular}} & \textbf{\begin{tabular}[c]{@{}c@{}}Move \\ Bowl\end{tabular}}& \textbf{\begin{tabular}[c]{@{}c@{}}Move \\ Veggies\end{tabular}} & \textbf{Average}\\ \midrule
\textbf{\begin{tabular}[c]{@{}c@{}}Our method\end{tabular}} & 35\%                                                            & 30\%                                                          & 35\%                                                            & 40\%                                                           & 45\%   & 35\%   & 37\%
\\
\textbf{\begin{tabular}[c]{@{}c@{}}Our (less data)\end{tabular}} & 10\%                                                            & 5\%                                                          & 10\%                                                            & 15\%                                                           & 15\%   & 10\%   & 11\%
\\
\textbf{\begin{tabular}[c]{@{}c@{}}3D Scene Flow\end{tabular}} & 15\%                                                            & 15\%                                                          & 5\%                                                            & 10\%                                                           & 15\%    & 10\%   & 12\% \\\bottomrule
\end{tabular}
\label{tb:goalconditional}
\end{table}
\setlength{\tabcolsep}{4pt}
\begin{table}[h!]
\vspace*{-0.3cm}
\centering
\caption{\footnotesize Given a scene with bottom drawer half open, and top drawer closed, we evaluate 10 trials of the unconditioned model, and count trials (out of 10) that reach a certain final configurations . We compare this to the goal-conditioned model that gets conditioned on a goal corresponding to a final configuration (10 trials per goal).}
\begin{tabular}{@{}cccc@{}}
\toprule
\textbf{Final Config Reached./ Goal}                    & \textbf{Top Open}    & \textbf{Bottom Open} & \textbf{Bottom Close} \\ \midrule
\textbf{Unconditioned  (10 trials overall)} & 1/10 & 3/10  & 2/10   \\ \midrule
\textbf{Goal-conditioned   (10 trials each)} & 3/10 & 4/10 &  4/10  \\ 
\bottomrule
\end{tabular}
\label{tb:ungoal}
\vspace*{-0.3cm}
\end{table}
\subsection{Analysis of failures}

For the unconditioned model, we observe two primary failure modes: when the robot fails to make contact with the object in the initial time-step (about 40\% of all failures) and when it does make contact, but the resulting motion is not feasible and the robot gets stuck (about 60\% of all failures). The second failure mode corresponds to a predicted trajectory that doesn't align with the affordances of the respective object, for example trying to pull a door outwards, or pressing against the side of a bowl resulting in the bowl toppling on the table. In addition to the above failures, for the goal conditioned model we observe a third failure mode, where the executed trajectory is plausible (for example a closer drawer is opened) but doesn't correspond to the specified goal (for example the goal shows the top drawer opened, whereas the executed trajectory opens the bottom drawer). This corresponds to about 40\% failures of the model. 

\section{Discussion and Conclusion}
In this paper we developed an approach for predicting plausible action trajectories from passive human videos, for zero-shot robot interaction given a scene with objects. After learning a model to predict agent-agnostic action trajectories from human videos on the web, we transformed the predictions to the robot's embodiment, and executed the motions zero-shot in a robot workspace without fine-tuning on any in-lab data. For everyday objects like drawers, toaster ovens, doors, and fruit bowls we observed that the predicted trajectories performed plausible interactions with a success rate of around 50\%, for example opening a drawer that is closed, and moving a bowl on the table. We further developed a goal-conditioned version of the model that conditions on an initial image of a scene, and a goal image, trained again from egocentric human videos on the internet where the goal image is a final image of a video clip. When the goal-conditioned model is deployed in the robot workspace, we observed around 40\% success rate in reaching a specified goal, for example a door fully open, from an initial scene, for example a door half open. We believe our approach for scene-conditioned action trajectory generation would help understand the limits of extracting action representations from passive human videos alone, such that they are useful for robot manipulation zero-shot without any fine-tuning. 

\newpage
\clearpage
\bibliography{references}

\newpage
\clearpage

\section{Appendix}

\subsection{Experiment details}

For the experiments, we consider five different everyday objects shown in Fig.~\ref{fig:scene_objects}, with a total of ten different semantic tasks possible with these objects. The possible tasks are moving veggies on the chopping board, opening the door, closing the door, opening the top drawer, closing the top drawer, opening the bottom drawer, closing the bottom drawer, opening the toaster oven, closing the toaster oven, and moving bowl of fruits across table. The objects and tasks cover different types of motions like sliding with a tool (chopping board), rotation about vertical hinge (door), rotation about horizontal hinge (toaster), and pushing (bowl of fruits). For the chopping board task, we constrain the motion to be within the plane of the table with vertical motion restricted to within 1 cm from the table (so that the spatula doesn't hit the table and fall off the gripper). Results with these different possible behaviors aim to show the generality of the proposed approach in generating both unconditional (when task is not known apriori) and goal-conditioned (when task is specified with goal image) behaviors.

We consider a single IntelRealsene camera in the scene, with RGB image observation that is used as input to our prediction model. For the goal-conditioned setting, the goal image is obtained with the same camera. The camera is fixed and is calibrated. The robot is a 7DOF Franka Emika Panda arm, operated with an IK controller. The base of the Franka is fixed, and its location is known. The End-Effector is a two-finger adaptive Robotiq gripper.

For the baselines, we consider a scene flow~\cite{sceneflow,raft3d} baseline that uses RAFT3D~\cite{raft3d} for predicting scene flow field between the initial and goal images, and uses the dominant flow direction to guide the motion of the robot. This baseline uses depth image (RGBD from the same camera) to compute scene flow and so requires more information that our method. To test the importance of training across diverse data, we compare against a version of our method that is trained on only 30\% of the total training data, with everything else kept the same for training and evaluation.

\subsection{Additional details on the Approach}
The model architectures are described in section~\ref{sec:model}. Fig.~\ref{fig:architecture} shows the un-conditioned model. The goal-conditioned model is very similar, with an additional goal image provided as input, with features $f_g$. This is encoded bu the Transformer encoder into $z_g^e$. There is an additional positional embedding dimension in the encoding to disambiguate between the initial image embedding and the goal image embedding. The rest of the architecture is same as that described in section~\ref{sec:model}. For the transformer, the embedding dimension is 512, and dropout rate is 0.1 for all blocks. The CVAE networks for encoder and decoder are implemented as 2-layer MLPs. The architecture is auto-regressive such that prior actions are used as input to predict the current action. As is standard in auto-regressive generation, during training, we feed in the previous ground-truth actions at each time-step, and during inference, we feed in the previous predicted actions.

After training the overall hand pose prediction model, $p_\psi( a_{1:T}|o_1)$ with diverse internet videos, we deploy it directly for robot manipulation tasks in the lab. The robot sees an image of the scene through a fixed camera, and optionally receives a goal-image which is input to the prediction model. In order to use the actions predicted by the model $a_{1:T}$ for moving the robot, we need to transform them to the world coordinate frame of the robot, and considering each action $a_t$ as an end-effector target pose, use a low-level controller for executing the respective motions. 

 The camera in the scene is calibrated, so the intrinsic matrix $I$ and the extrinsic matrix $[R,T]$ are known. The world coordinates are located at the base of the robot (robot base is at same height as the table top) and the height of the table top from the camera is  known and approximately constant. Given scene from the camera $o_1$, the model predicts delta actions $a_{1:T}$ which we convert to absolute actions (described in section~\ref{sec:model}), and transform the actions from the camera frame to the world frame of the robot through inverse projection transformation. The prediction horizon is $T=7$ for our experiments. After obtaining the world coordinates of the action sequence $\{(X_t,Y_t,Z_t,\alpha_t,\beta_t,\gamma_t)\}_{t=1}^T$, we use an IK controller to execute the corresponding motion for bringing the end-effector to the desired position and orientation and each time-step. The IK controller has an error threshold of 10\% for position and 20\% for orientation so that the robot doens't get stuck trying to reach a difficult predicted pose.
\end{document}